\PassOptionsToPackage{x11names}{xcolor}
\documentclass[letterpaper, 10 pt, conference]{ieeeconf}  

\IEEEoverridecommandlockouts                              
\usepackage[utf8]{inputenc}
\usepackage[T1]{fontenc}
\usepackage{graphicx}
\usepackage{caption}
\usepackage{amssymb}
\usepackage{comment}
\usepackage{stfloats}
\usepackage{tikz}
\usepackage{adjustbox}
\usepackage{subcaption}
\usetikzlibrary{calc}
\usepackage{grfext}
\usepackage{multirow} 
\usepackage{booktabs}
\usepackage{array}
\makeatletter
\let\labelindent\relax
\makeatother
\usepackage{enumitem}
\usepackage[x11names]{xcolor}
\usepackage{pgfplots}
\pgfplotsset{compat=1.18}
\usepackage{amsmath}
\definecolor{graybox}{HTML}{808080}
\definecolor{bluebox}{HTML}{3399FF}
\definecolor{greenbox}{HTML}{97D077}
\definecolor{orangebox}{HTML}{D79B00}

\makeatletter
\let\NAT@parse\undefined
\makeatother

\definecolor{rblue}{rgb}{0,0.5,1}
\definecolor{awesome}{rgb}{1.0, 0.13, 0.32}
\definecolor{hollywoodcerise}{rgb}{0.96, 0.0, 0.63}
\definecolor{lasallegreen}{rgb}{0.03, 0.47, 0.19}
\definecolor{hanpurple}{rgb}{0.32, 0.09, 0.98}
\definecolor{green(pigment)}{rgb}{0.0, 0.65, 0.31}
\usepackage[pagebackref=false,breaklinks=true,colorlinks,bookmarks=false]{hyperref}
\hypersetup{colorlinks,linkcolor={red},citecolor={hanpurple},urlcolor={magenta}}  

\title{\LARGE \bf
Visually Grounded Narratives: Reducing Cognitive Burden in Researcher-Participant Interaction
}

\author{Runtong Wu$^{1,*}$, Jiayao Song$^{3,*}$, Fei Teng$^{2,*}$, Xianhao Ren$^{4}$, Yuyan Gao$^{5,{\dag}}$, Kailun Yang$^{2}$
\thanks{$^{1}$The author is with the School of Mathematics, Hunan University, China.}
\thanks{$^{2}$The authors are with the School of Artificial Intelligence and Robotics, Hunan University, China.}
\thanks{$^{3}$The author is with International Business School, Henan University of Economics and Law, China.}
\thanks{$^{4}$The author is with Science \& Technology College, University of Lorraine, France.}
\thanks{$^{5}$The author is with the Institute for Language Education, the University of Edinburgh, UK.}
\thanks{$^{*}$Equal contribution.}
\thanks{$^{\dag}$Corresponding authors: Yuyan Gao (email: Y.Gao-120@sms.ed.ac.uk).}
}

\begin{document}

\IEEEaftertitletext{%
  \vspace{-1em}
  \begin{flushleft}
    \addtocounter{figure}{1}
    \hypertarget{fig:pipeline}{}
    \includegraphics[width=\textwidth]{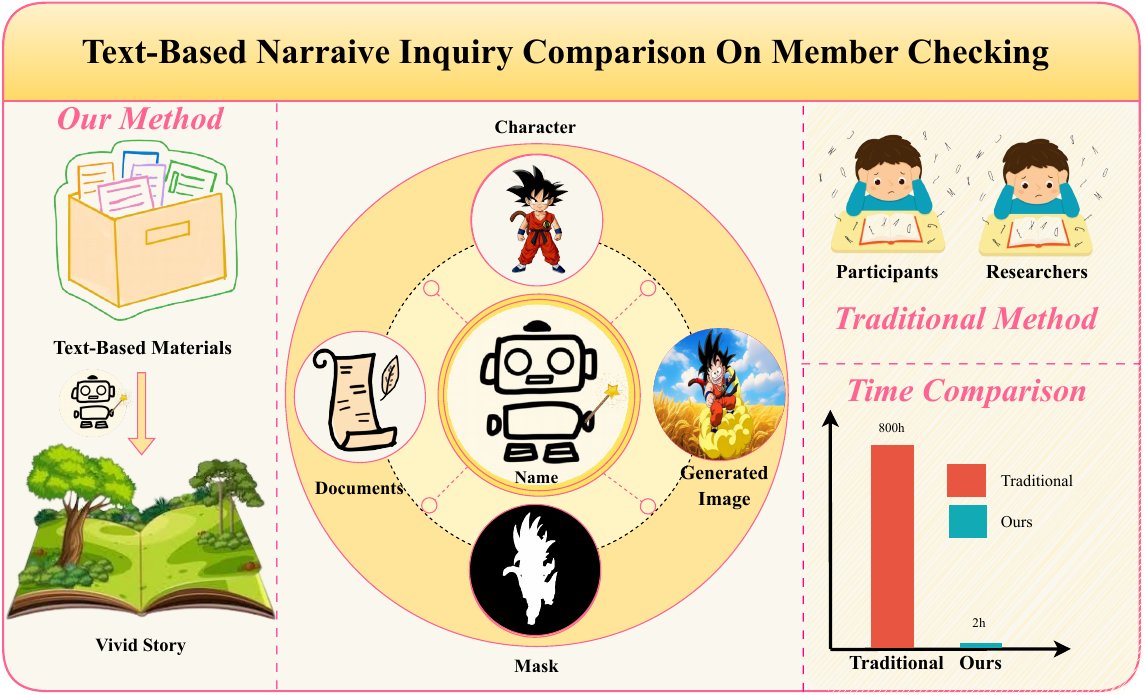}
    \\[0.5em]
    \small\textbf{Fig. \thefigure.} Comparison of member checking between our method and traditional practices. Our approach utilizes a reference prompt, reference image, input prompt, and spatial mask to generate character-coherent story sequences with precise character positioning. During member checking, our model can automatically transform textual materials into consistent, visually grounded character images. Our method completes the entire pipeline in 2 hours, with minimal workload during image-based checking, compared to up to 800 hours required by traditional approaches, where researchers have to analyze large volumes of data, and participants are asked to review extensive processed content to validate the researchers' interpretations under a significant cognitive burden.
  \end{flushleft}
  \vspace{1em}
}

\maketitle
\thispagestyle{empty}
\pagestyle{empty}

\maketitle
\thispagestyle{empty}
\pagestyle{empty}

\begin{abstract}

Narrative inquiry has been one of the prominent application domains for the analysis of human experience, aiming to know more about the complexity of human society. However, researchers are often required to transform various forms of data into coherent hand-drafted narratives in storied form throughout narrative analysis, which brings an immense burden of data analysis. Participants, too, are expected to engage in member checking and presentation of these narrative products, which involves reviewing and responding to large volumes of documents. 
Given the dual burden and the need for more efficient and participant-friendly approaches to narrative making and representation, we made a first attempt: (i) a new paradigm is proposed, NAME, as the initial attempt to push the field of narrative inquiry. Name is able to transfer research documents into coherent story images, alleviating the cognitive burden of interpreting extensive text-based materials during member checking for both researchers and participants. (ii) We develop an actor location and shape module to facilitate plausible image generation.
(iii) We have designed a set of robust evaluation metrics comprising three key dimensions to objectively measure the perceptual quality and narrative consistency of generated characters. 
Our approach consistently demonstrates \textit{state-of-the-art} performance across different data partitioning schemes. Remarkably, while the baseline relies on the full $100\%$ of the available data, our method requires only $0.96\%$ yet still reduces the FID score from $195$ to $152$. Under identical data volumes, our method delivers substantial improvements: for the $70{:}30$ split, the FID score decreases from $175$ to $152$, and for the $95{:}5$ split, it is nearly halved from $96$ to $49$. Furthermore, the proposed model achieves a score of $3.62$ on the newly introduced metric, surpassing the baseline score of $2.66$. Beyond quantitative gains, our work enhances the efficiency of member checking and reimagines the interaction between narrative inquiry researchers and participants' stories---shifting from labor-intensive textual analysis toward a more accessible, visually grounded mode of inquiry that both respects the analytical expertise of researchers and safeguards the well-being of participants. The source code will be released.

\end{abstract}

\section{Introduction} 

Narrative inquiry is a methodology that centers on human experiences. It seeks to understand individuals' inner world by making sense of their experiences through narrative structures, which offer rich insights into the intricacies and complexity of human phenomena that are of our research interest while keeping a holistic view  \cite{booktt} \cite{Polkinghorne1995NarrativeCI}. Having been widely used in fields such as education, sociology, and healthcare, this methodology has gained particular popularity for exploring sensitive issues like psychological trauma, identity formation, and childhood development \cite{https://doi.org/10.1016/j.adolescence.2019.08.007} \cite{Hall2011} \cite{doi:10.1177/0959354300104005} \cite{Shiner2021} \cite{Niiranen2024}. By foregrounding temporality, context, and subjectivity, narrative inquiry offers a methodological framework for examining experiences that, while not easily quantifiable, are nonetheless crucial for comprehending the complexities of the social world \cite{Misztal2020} \cite{Gjessing2023}.

However, narrative inquiry is a highly labor-intensive endeavor \cite{chase2008narrative} \cite{articleyu}, primarily reflected in the following two aspects. (i) Transformation of data that are diverse in terms of both form and content from various sources -such as interviews, field notes, or audio/visual recordings-into coherent and temporally organized narrative texts. This process typically requires researchers to manually synthesize fragmented material into structured stories, which is both intellectually demanding and time-consuming, especially in large-scale projects \cite{Nasheeda2019} \cite{Ayton2023} \cite{Phillippi2017}. (ii) In addition to the workload faced by researchers, the process of member checking-which invites participants to verify or reflect on the narratives constructed from their accounts-can also pose significant
 challenges for participants themselves. One concern is cognitive and practical burden: participants are often asked to read and assess long-form textual narratives, which may be difficult for those with limited time, literacy, or familiarity with academic discourse. This can lead to fatigue, disengagement, or even anxiety \cite{Caretta2019} \cite{GOLDBLATT2011389}. A second, and more ethically pressing, concern is the risk of psychological distress. Member checking may require participants to revisit emotionally charged or traumatic memories as they review and validate the researcher's interpretation. A second concern is the potential risk of psychological distress. If the images used for member checking are of poor quality or visually incoherent, they may cause confusion or discomfort. Asking participants to review such images may inadvertently expose them to disturbing content, potentially eliciting emotional distress \cite{article} \cite{Sanjari2014} \cite{sexes6020028}.

With the widespread adoption of Transformer \cite{Vaswani2017Attention} architectures and significant advancements in computational power, generative artificial intelligence has made remarkable progress in both content coherence and quality \cite{Bommasani2021} \cite{Villegas2019} \cite{Singer2022}. For instance, large language models can assist with coding, question answering, and translation tasks \cite{OpenAI2023} \cite{Touvron2023}. In the field of background music generation, generative AI enables the creation of contextually appropriate music tailored to specific scenes \cite{li2024diffbgmdiffusionmodelvideo} \cite{polyffusion2023}. This study constitutes the first attempt to incorporate generative artificial intelligence into narrative inquiry, opening up new possibilities for creative endeavors, offering a novel perspective on controllable generation, and broadening the potential applications of generative models. Our model significantly reduces the time required for member checking from approximately 800 hours, as seen in traditional methods, to just 2 hours. This reduction translates into substantial savings in human and material resources, including labor costs, scheduling efforts, and communication overhead. While current Text-to-Image (T2I) models have achieved remarkable advancements in image quality, they often neglect the psychological states of participants, which can adversely affect their mental well-being \cite{McTeague2010} \cite{OKearney2006} \cite{Hayes2012}. To address this limitation, our study prioritized the reduction of participants' cognitive and emotional burden by simplifying language, ensuring materials were accessible and non-threatening, and incorporating supportive visual aids. These design choices were integral to safeguarding participants' psychological well-being and fostering a respectful, emotionally secure research environment-an approach particularly well-aligned with the principles of narrative inquiry.

Our method substantially reduces the labor costs associated with member checking in narrative inquiry, thereby contributing significantly to the efficiency of qualitative research validation. An overview of the member checking time consumption and execution time comparison among the proposed and traditional methods is shown in \hyperlink{fig:pipeline}{Fig.~\thefigure}. The main contributions can be summarized as follows: 

(i) We proposed a paradigm that reduces the interpretive load on participants by transforming narrative materials into more accessible multimodal forms.  By leveraging generative models to convert complex textual narratives into visual representations, we aim to support intuitive understanding while preserving narrative coherence and nuance. 

(ii) We proposed a controllable generation module that enables precise manipulation of character positioning within generated images, a feature essential for maintaining narrative clarity by visually reinforcing roles, relationships, and scene structure. By allowing researchers to guide the spatial semantics of generation, our module ensures that outputs remain both semantically accurate and emotionally considerate. Additionally, we modified the existing dataset and provided a new benchmark.

(iii) We develop a comprehensive set of evaluation metrics, structured around three core dimensions, to objectively assess the perceptual quality and narrative coherence of generated characters, as well as to reflect the cognitive and interpretive burdens experienced by both participants and researchers during the member-checking process. 

\section{Related Work}
 
\subsection{Narrative Inquiry }
In recent years, the social sciences have undergone a `narrative turn', prompted by the growing recognition that research approaches modeled on the natural sciences are inherently limited when applied to human problems \cite{barkhuizen2025narrative}. As a consequence, narrative inquiry has emerged as an `alternative paradigm for social researchers \cite{lieblich1998narrative}.

Narrative inquiry, which began in literary studies, has gradually developed into a multidisciplinary approach. It is now widely applied across various fields, including psychology, education, medicine, sociology, anthropology, economics, history, and sociolinguistics \cite{lieblich1998narrative} \cite{riessman2008narrative}. At the heart of narrative inquiry lies an interest in the ways individuals use narratives to interpret their lived experiences, especially in contexts that require an understanding of events from participants' own viewpoints. \cite{wells2011introduction}.

Narrative is often defined in connection with an event involving a change of state, which is conveyed in discourse through a process statement expressed in the mode of `Do' or `Happen'. Such a change of state is also considered one of the fundamental components of a story \cite{prince2003dictionary}.

Although narrative inquiry adopts diverse approaches, it commonly treats stories as the primary data source and focuses on comprehensive analyses of entire accounts - integrating content, structure, performance, and context - rather than dissecting them into separate thematic elements \cite{prince2003dictionary}. For ethical reasons, narrative inquiry researchers are suggested to ask participants to review and comment on these accounts or the data to be included in a study during analysis (i.e., member checking). 

In narrative inquiry, we introduce a controllable framework that preserves character consistency and spatial positioning. By transforming textual narratives into vivid visual representations, our approach enhances data processing efficiency during the member checking process.

\subsection{Text-to-Image Generation} 

Text-to-image generation has long been a prominent research topic, aiming to translate natural language descriptions into corresponding visual content by learning cross-modal correspondences from large-scale multimodal datasets. Over the years, three primary frameworks have shaped the development of this field: Generative Adversarial Networks (GANs) \cite{Goodfellow2020GAN}, auto-regressive models, and diffusion models. As one of the earliest and most influential approaches, GANs generate visually compelling images through adversarial training between a generator and a discriminator. Several GAN-based methods have demonstrated strong performance in synthesizing images from text \cite{xu2017attnganfinegrainedtextimage} \cite{zhang2017stackgantextphotorealisticimage} \cite{ zhang2018stackganrealisticimagesynthesis}. Although their influence had waned with the emergence of newer paradigms, a number of subsequent works have brought renewed attention to GAN-based methods by proposing more effective and streamlined architectures \cite{tao2022dfgansimpleeffectivebaseline} \cite{ tan2022drgandistributionregularizationtexttoimage}. Alongside GANs, Auto-regressive models such as those presented in \cite{Ramesh2021ZeroShot} \cite{ ding2021cogviewmasteringtexttoimagegeneration} \cite{ ding2022cogview2fasterbettertexttoimage}, leverage the Transformer architecture \cite{Vaswani2017Attention} to facilitate stable training and produce high-fidelity image outputs. Another paradigm that plays a dominant role in text-to-image generation is diffusion. Notably, Stable Diffusion \cite{rombach2021highresolution}, which operates in a latent space, significantly reduces computational cost while maintaining high visual fidelity. These models excel at generating images with fine-grained semantic details and have set new standards in Text-to-image generation. Such capabilities have led some studies to extend its application into the field of story generation \cite{tao2024storyimagerunifiedefficientframework} \cite{he2025dreamstoryopendomainstoryvisualization} \cite{chen2024mangadiffusion} \cite{wu2025diffsenseibridgingmultimodalllms}, demonstrating its potential in generating multimodal narrative content. Building upon diffusion models, our work represents a first attempt to apply this approach within the field of narrative inquiry. Drawing on the foundational definition of a story-as involving a change of state-we strictly adhere to this definition: our primary objective is to integrate diffusion-based techniques into narrative inquiry in a way that remains faithful to its theoretical underpinnings. Our model significantly reduces the cognitive and interpretive burden on both participants and researchers during member checking, thereby saving valuable time and enhancing overall efficiency.

\subsection{Diffusion models} 

Diffusion models have recently emerged as a powerful generative framework that synthesizes high-quality images through an iterative denoising process starting from Gaussian noise. Since the introduction of DDPM \cite{Ho2020DDPM}, the field has rapidly expanded, with works such as DDIM \cite{song2022denoisingdiffusionimplicitmodels} accelerating the sampling process while maintaining generation quality. Conditional diffusion models have gained prominence, with classifier-guided and classifier-free guidance allowing for flexible control over generation via modalities like text or images. Latent Diffusion Models (LDM) incorporate Variational Autoencoders (VAE) \cite{kingma2022autoencodingvariationalbayes} to shift the denoising process into the latent space, significantly reducing computational cost. Based on this, Stable Diffusion \cite{rombach2021highresolution} integrates CLIP, attention mechanisms, and LDM to synthesize high-fidelity images. These techniques have been widely applied beyond static image generation, including in music \cite{polyffusion2023, li2024diffbgmdiffusionmodelvideo, mittal2021symbolicdiffusion}, video \cite{ho2022imagenvideohighdefinition, ho2022videodiffusionmodels}, Audio \cite{kong2021diffwaveversatilediffusionmodel} \cite{božić2024surveydeeplearningaudio} \cite{chen2021wavegrad2iterativerefinement} \cite{liu2022diffganttshighfidelityefficienttexttospeech} \cite{huang-etal-2023-fastdiff} and medical imaging \cite{pub.1156959064}, showcasing their versatility.

In addition, a number of diffusion-based controllable generation methods have been adopted to enhance structural guidance during inference. ControlNet \cite{zhang2023adding} enables precise conditioning by injecting auxiliary inputs-such as depth maps, edge detections, and sketches-into the generation process. GLIGEN \cite{li2023gligenopensetgroundedtexttoimage} employs bounding box annotations to explicitly control the spatial layout of generated objects. T2I-Adapter \cite{mou2023t2iadapterlearningadaptersdig} introduces lightweight adapter modules that can be seamlessly integrated into existing diffusion pipelines, offering controllability without the need for extensive retraining. Furthermore, Various domain-specific methods, addressing controllable generation, visual quality optimization, and diverse applications, have been proposed and empirically validated within the diffusion model framework. \cite{zhou2025attentiondistillationunifiedapproach} \cite{shen2023learningglobalawarekernelimage} \cite{he2025conceptrolconceptcontrolzeroshot} \cite{nam2025visualpersonafoundationmodel} \cite{wang2025designdiffusionhighqualitytexttodesignimage} \cite{liu2025onepromptonestory}. These advancements demonstrate the growing sophistication and adaptability of diffusion models, motivating researchers' adoption of diffusion-based approaches to address the challenges brought by the large volume of data in narrative inquiry. Compared to previous work, which primarily focuses on character consistency and image quality, our approach introduces spatial and geometric constraints through the use of masks. This allows for more coherent spatial positioning and geometrical structure of the generated characters. By doing so, we mitigate the likelihood of logically implausible generations and reduce the risk of producing content that may be emotionally unsettling or discomforting for participants. For both researchers and participants involved in member checking phase of the narrative inquiry, these models offer improved accuracy in interpreting textual data, reduced cognitive and interpretive demands, and greater efficiency in terms of labor and time expenditure.

\section{Methodology}

In this section, we describe the position and shape control module, along with a simple yet effective component designed to enhance model performance.

\subsection{\textit{Problem Formulation}}

In narrative inquiry, member checking is a widely adopted strategy for ensuring the trustworthiness of qualitative interpretations. However, this process can impose considerable cognitive demands. For researchers, these challenges often arise during the analysis and synthesis of large volumes of textual data. For participants, reviewing and validating lengthy narrative accounts can be overwhelming and mentally exhausting. An ideal approach would therefore aim to reduce the cognitive load associated with textual interpretation and streamline qualitative research workflows. Incorporating images into the member checking process offers one such solution by making complex narrative data more accessible, concrete, and easier to engage with. 

Prior research in cognitive psychology and visual studies suggests that replacing or supplementing text with images can substantially lower the mental effort required for information processing. By tapping into dual-coding mechanisms, visual stimuli offload the burden on verbal working memory \cite{10.1207} \cite{sweller1998cognitive} and allow viewers to process meaning through complementary channels, thereby reducing cognitive load and enhancing overall comprehension \cite{paivio2013imagery} \cite{Mayer_2009} \cite{069d6b7c-5317-393c-a548-0192069176ab}. Images-whether photographs, illustrations, or data visualizations-serve as concrete anchors that transform abstract or fragmented textual descriptions into coherent, retrievable mental structures. In addition to facilitating more accurate recall and richer narrative construction, visual materials boost engagement by invoking emotional resonance and personal associations, making participants more invested in the validation process \cite{doi:10.1177/001139286034003006}. They also promote inclusivity-helping individuals with diverse literacy levels or language backgrounds to grasp content more readily-and invite reflexivity by provoking deeper self-reflection on identity and experience \cite{alma9924461324102466} \cite{alma991010422409706011}. Finally, visuals can accelerate pattern recognition and comparative analysis \cite{articlel}, enabling quicker feedback loops during member checking and strengthening the rigor of participatory research \cite{riessman2008narrative}.

Recent advances in the story synthesis field \cite{he2025dreamstoryopendomainstoryvisualization} \cite{tao2024storyimagerunifiedefficientframework} \cite{Liu_2024_CVPR} have focused on producing high-quality, visually coherent images with consistent character representations. However, these models fail to consider participants' psychological responses. The generation of inappropriate or incongruent images may elicit discomfort or cognitive dissonance, potentially leading to adverse emotional effects \cite{doi:10.1068/p5814}. 

\begin{figure*}[htbp]
  \centering
  \includegraphics[
    width=1\textwidth
  ]{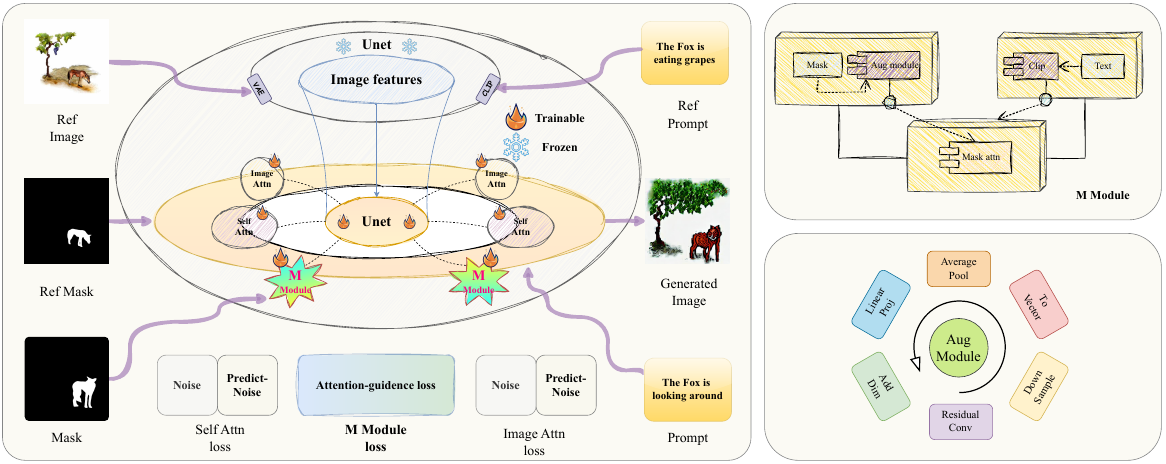}
  \caption{Model Overview. Our model takes a reference image-mask-text triplet and a target text-mask pair as input conditions, and generates a story through regressive generation with consistent character identity and spatial positioning. To ensure accurate self-attention and image-attention mechanisms, we compute the added noise and predicted noise as part of the loss function. For character-specific positional generation, the masks define desired attention regions, and the loss penalizes deviations from these regions accordingly. 
  The $M$ module processes the input masks via an Aug module and the text via CLIP, then integrates both modalities. In the Aug module, the mask first passes through a residual convolutional block, followed by two downsampling operations and a vectorization step. Finally, average pooling and a linear projection are applied to align the mask representation with the text features.}
  \label{fig:system-overview}
\end{figure*}

To address this limitation, we adopt StoryGen \cite{Liu_2024_CVPR} as our base model and introduce modifications to its cross-attention mechanism, enabling controllable character positioning while preserving character coherence across frames. Specifically, the model generates the current frame \(I_{k}\) by conditioning on the current spatial mask \(M_{k}\), the textual prompt \(T_{k}\), and preceding text-image-mask pairs. The overall generation process is formalized as equation~\eqref{eq:ik-definition}:
\begin{equation}
I_{k} := \phi(T_{k}, M_{k}, (I_{<k}, T_{<k}, M_{<k})),
\label{eq:ik-definition}
\end{equation} here \(\phi()\) refers to our model. The story \(\{ I_{1}, I_{2}, \dots, I_{n} \}\) could be visualized by step-by-step inference. Our model overview is illustrated in Fig.~\ref{fig:system-overview}.

Our model is able to: (i) generating images on any given storyline; (ii) synthesizing process could be extended to any characters that have not yet been introduced; and (iii) controlling the main character's location and shape.

\subsection{\textit{Latent Diffusion Models}}

Latent Diffusion Models (LDMs) are a class of generative models that perform the diffusion process in a learned latent space instead of the original pixel space \cite{rombach2021highresolution}. Compared to standard diffusion models that operate in high-dimensional space, LDMs achieve significant improvements in computational efficiency while maintaining high sample quality.

An encoder \( E \) first maps the input image \( x \in \mathbb{R}^{H \times W \times C} \) into a lower-dimensional latent representation \( z = E(x) \). The forward diffusion process is then applied to \( z \), where Gaussian noise is gradually added over \( T \) steps:
\begin{equation}
  q(z_t \mid z_{t-1}) = \mathcal{N}\left(z_t; \sqrt{1 - \beta_t}\, z_{t-1},\, \beta_t I \right),
\end{equation}
with a fixed variance schedule \( \{\beta_t\}_{t=1}^T \). This produces a sequence of increasingly noisy latent variables \( z_1, z_2, \ldots, z_T \). A neural network \( \epsilon_\theta(z_t, t) \), typically a time-conditional U-Net, is trained to estimate the added noise at each step. The training objective minimizes the prediction error of the noise:
\begin{equation}
  \mathcal{L}_{\mathrm{LDM}} =
  E_{z = E(x),\, \epsilon \sim \mathcal{N}(0,1),\, t} \left\|
  \epsilon - \epsilon_\theta(z_t, t)
  \right\|_2^2,
\end{equation}
where \( z_t \) is a noisy version of \( z \) at timestep \( t \), and \( \epsilon \) is the sampled noise. At inference time, starting from a sampled latent noise \( z_T \sim \mathcal{N}(0, I) \), the model iteratively denoises to obtain \( z_0 \), which is then decoded by \( D \) to reconstruct the image: \( \hat{x} = D(z_0) \). 

Latent diffusion models (LDMs) can incorporate external information (e.g., text embeddings or class labels) by introducing conditioning vectors into the denoising network, typically via concatenation or cross-attention mechanisms. These conditioning strategies improve controllability and strengthen the semantic alignment between inputs and outputs, thereby enabling a broad spectrum of applications such as conditional generation and inpainting tasks.

\subsection{\textit{Base model overview}}
StoryGen is developed on the foundation of a pre-trained stable diffusion model and incorporates a novel cross-attention module. By leveraging both previous text-image pairs and the current prompt, it facilitates the production of images that maintain character consistency. 
To be specific, the model initially adds noise to preceding frames to extract features, applies pre-trained SDM to denoise under the guidance of corresponding text, and selects features after the self-attention layer in Unet blocks as context features. The process is shown in equation~\eqref{stage1}:
\begin{equation}
F = [\varphi_{\mathrm{SDM}}({I}_1, \varphi_{\mathrm{CLIP}}(T_1)), \dots, \varphi_{\mathrm{SDM}}({I}_{k-1}, \varphi_{\mathrm{CLIP}}(T_{k-1}))].
\label{stage1}
\end{equation} Subsequently, a new cross-attention was added after the original two attention layers. For previous features,the query (\(Q_{p}\)) is derived from noised latent and key (\(K_{p}\)) and value (\(V_{p}\)) are derived from F; for current features,the query (\(Q_{T}\)) is also derived from noised latent but key (\(K_{T}\)) and value (\(V_{T}\)) are both derived from current text. The output could be formulated as equation~\eqref{stage2}:
\begin{equation}
O = \mathrm{Softmax}\left( \frac{Q_p (K_p)^\top}{\sqrt{d}} \right) V_p + \mathrm{Softmax}\left( \frac{Q_T (K_T)^\top}{\sqrt{d}} \right) V_T.
\label{stage2}
\end{equation} Ultimately, as for image generation, they adapted a novel classifier-free guidance term. Two guidance scales $\omega_v$ and $\omega_T$ are used for the visual condition and the text condition. The final noise for inference $\bar{\epsilon_\theta}$
and UNet-predicted noise $\epsilon_\theta$ relationship could be formulated as equation~\eqref{control}:

\begin{align}
\bar{\epsilon_\theta}(x_t, t, \mathcal{C}_V, \mathcal{C}_T) &= \epsilon_\theta(x_t, t, \emptyset, \emptyset) \nonumber \\
&\quad + w_v \left( \epsilon_\theta(x_t, t, \mathcal{C}_V, \emptyset) - \epsilon_\theta(x_t, t, \emptyset, \emptyset) \right) \nonumber \\
&\quad + w_t \left( \epsilon_\theta(x_t, t, \mathcal{C}_V, \mathcal{C}_T) - \epsilon_\theta(x_t, t, \mathcal{C}_V, \emptyset) \right).
\label{control}
\end{align} Given its outstanding performance on character consistency, we choose StoryGen as our baseline and intend to achieve controllable image generation with respect to character location and shape.

\subsection{\textit{Model}}
In addition to the baseline inputs, our model incorporates two additional components: a reference image mask designed to enhance model performance, and a positional mask that guides character placement while further improving generation quality. In the following sections, we elaborate on two key modules-the Masked Cross-Attention Layer and the Augmentation Module-which are integral to these enhancements.

\textbf{Mask Cross-attention layer}
Inspired by MAG \cite{endoTVC2023}, we also chose to edit the noised maps in the mask cross-attention layer to realize the controllable character position synthesis. Specifically, we changed the estimated cross-attention map with a constant map, drawing $1/sum$ at the target region and zero at others, where $sum$ represents the number of words in the prompt. Loss function can be formulated as equation~\eqref{MAG}:
\begin{equation}
L = \sum_{i=1}^{N} \sum_{w \in C_i} \sum_{p \in S_i} M_{p,w} + \lambda \sum_{i=1}^{N} \sum_{w \in C_i} \sum_{p \in \bar{S}_i} M_{p,w},
\label{MAG}
\end{equation} where $M_{p,w}$is a value of a cross-attention map $M$ for the word $w$ at pixel $p$, and $\lambda$ is a balancing weight.

While editing the noise map can achieve controllable character positioning, we noticed that MAG incurs a bit drop in performance metrics compared to its base model. However, our objective extends beyond achieving mask-based control-we also aim to improve overall model performance. We assumed that this performance degradation may arise from two primary reasons:
(i) MAG applies noise map editing only during inference, which may lead the model to prioritize spatial arrangement over image fidelity.
(ii) Their method involves directly assigning values within the noise map, which compromises differentiability during backpropagation.

To address these concerns, we proposed corresponding solutions. To address the first concern (i), we incorporate this process into the training phase rather than the inference phase, as detailed in the Training section. As for Concern (ii), we reformulated the noise editing process to enhance differentiability. Instead of directly assigning values, we initialized a zero-map and then added the product of the mask and a normalized factor $1/sum$. This approach replaces direct value assignment with addition and multiplication operations, which are fully differentiable and thus more suitable for training. The revised process can be expressed as equation~\eqref{MAP}:
\begin{equation}
\text{MAP} = Z + \text{mask} \times \left( \frac{1}{\text{sum}} \right).
\label{MAP}
\end{equation} These modifications contribute to a more principled and flexible training paradigm. By shifting loss computation into the training stage, the model benefits from direct optimization signals that better reflect its generation objectives. Meanwhile, replacing non-differentiable assignments with smooth, learnable alternatives ensures uninterrupted gradient flow, which is essential for end-to-end learning. Together, these changes not only simplify implementation by reducing the need for ad hoc inference heuristics but also improve generalization by fostering tighter coupling between training dynamics and downstream performance.

\textbf{Augment Module} The idea was inspired by the text embedding fusion logic introduced by ControlNet \cite{zhang2023adding}, particularly its mechanism of conditionally integrating auxiliary embeddings. Given that CLIP encodes textual information into spatial representations, and that masks inherently contain spatial features, we experimented with fusing text embeddings and mask embeddings. It is important to note that this fusion in this module is intended solely to enhance image quality, rather than to control character positioning.

More specifically, it is the first to apply a residual convolutional block to the mask to prepare it for more efficient downsampling. This was followed by two downsampling operations to extract its spatial features. The resulting feature map was then passed through a to-vector module, which consisted of an average pooling layer and a Gaussian Error Linear Unit (GELU) activation function \cite{hendrycks2023gaussianerrorlinearunits}. Finally, we performed average pooling along the height and width dimensions and projected the resulting vector linearly to match the dimensionality of the text embeddings, then aligned it along the sequence length axis of the text representation. This formed the final mask embedding, which was then added to the original CLIP text embeddings. The overall pipeline is illustrated in Fig.~\ref{fig:system-overview}.

\subsection{\textit{Training}}
In this section, we detail our training strategy.   Our method consists of three stages: a single-frame pretraining stage, a character position fine-tuning stage, and a multi-frame fine-tuning stage. The training process is carefully designed to prevent the model from overfocusing on spatial layout at the expense of image quality or character consistency. To ensure that each module fulfills its intended role without interference, we adopted a staged training strategy. Specifically, we first trained the self-attention and image cross-attention layers to allow the model to learn how to generate visually coherent frames and maintain consistent character representations. Only after these foundational components were sufficiently optimized did we begin training the text-mask cross-attention layer. This ordering is motivated by findings from previous procedures, indicating that spatial layout control performs better when grounded in stable visual and character representations. Introducing positional supervision prematurely may cause the model to anchor layout patterns before it has a reliable understanding of character identity and appearance, which could hinder convergence or lead to degraded visual outputs. By deferring the training of the text-mask attention module, the model was allowed to first internalize what to generate, before learning where to place it. The detailed rationale and training strategies are presented in this section, while the effectiveness of our training strategy is further validated through ablation studies (A detailed explanation can be found in Section~\ref{Ablation}, Subsection~\ref{Attention layer}.). Specifically, during the single-frame pretraining stage, our model is built upon a standard Stable Diffusion Model (SDM), which is initially conditioned only on textual prompts. To improve the overall generation quality, we incorporated an additional mask embedding extracted from our augment module. The mask embedding does not serve a positional control function;  instead, it provides auxiliary information that enhances the model's capacity to synthesize visually rich single-frame outputs. In the subsequent phase, we fine-tune the image cross-attention layer to reinforce character consistency across frames.   We also continued to apply mask embeddings during this phase within both image feature extraction and image synthesis.   This decision is inspired by the architecture of our base model, which employs a U-Net to extract visual features from reference images.   Applying the injection of mask embeddings to the synthesis process helps enhance the integrity of character features. Finally, we began training the edited mask cross-attention module, which involves spatial layout control.   Here, the model integrates text embeddings and reference character features that encode positional constraints, enabling it to generate characters situated in specific, user-defined spatial contexts while maintaining character consistency.

\section{Experiments}

In this section, we provide a detailed description of our experimental setup and compare the images generated by our method with those from StoryGen. Additionally, we present the optimization results of our model in comparison with other models to validate the effectiveness and applicability of our proposed approach.

\subsection{\textit{Settings}}
We perform all training on a single NVIDIA GeForce RTX 4090 GPU. The learning rate and batch size are set to $1 \times 10^{-5}$ and $1$, respectively. The weighting coefficient $\lambda$ in the loss function is fixed at $0.5$.

To guide the model in learning both semantic and spatial representations, we adopt a \textbf{three-stage training scheme}:

\begin{itemize}
    \item \textbf{Stage 1}: We train the \textit{self-attention layers} for 15{,}000 epochs to enable the model to capture global semantic information from the input narratives.
    \item \textbf{Stage 2}: We then train the \textit{image cross-attention layers} for 50{,}000 epochs, refining and reinforcing the semantic understanding based on the visual context.
    \item \textbf{Stage 3}: Finally, we train the \textit{mask attention layers} for 25{,}000 epochs to enable precise control over character-level geometric positioning.
\end{itemize}

This staged approach ensures a progressive learning process, allowing the model to first establish strong semantic grounding before incorporating spatial control mechanisms.

\subsection{\textit{Automatic Evaluation results}}
\begin{figure*}[htbp]
  \centering
  \includegraphics[
    width=1\textwidth,
  ]{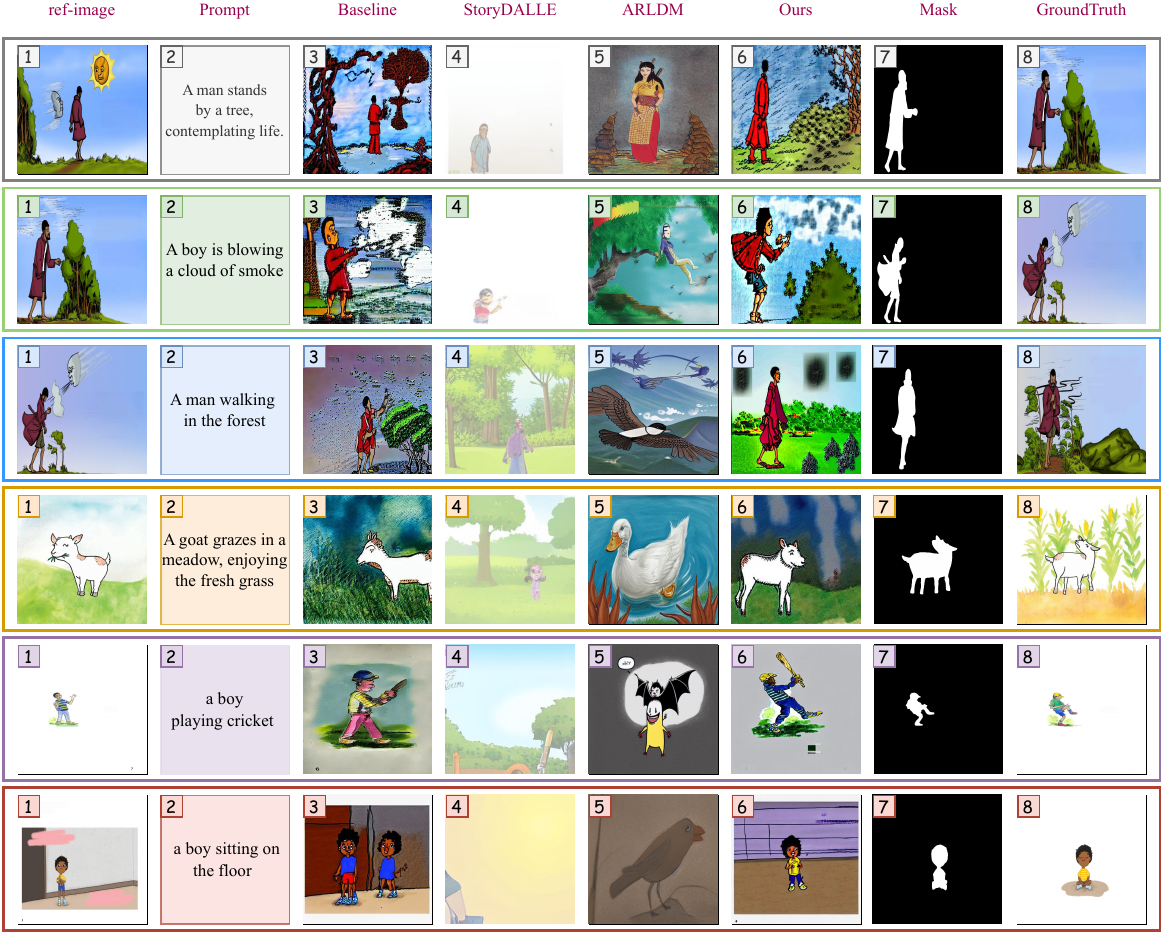}
  \caption{Visual Comparison on StoryGen (Baseline), StoryDALLE, ARLDM and NAME}
  \label{Comparison}
\end{figure*}

Since our approach involves selecting images containing only a single character to facilitate mask-based segmentation, we re-trained the model using our reconstructed dataset and divided our dataset into 70 percent for training and 30 percent for testing. For evaluation, we adopted the same metrics used in StoryGen, including Frechet Inception Distance score (FID) \cite{Seitzer2020FID}, CLIP image-image similarity (CLIP-I), and CLIP text-image similarity (CLIP-T). The results are summarized in Table~\ref{comparision_7:3} and the visulization results is showed in Fig~\ref{Comparison}:

Although our method does not achieve the best scores on all metrics, it demonstrates significant overall effectiveness. Specifically, we observe a substantial improvement in the FID score \cite{Seitzer2020FID}, indicating enhanced visual quality. On the CLIP-I and CLIP-T metrics, while our results are not the highest, the performance gap compared to the best-performing methods is marginal (less than 0.03). Unlike prior approaches that modify the cross-attention mechanism and often suffer from unstable performance, our method introduces character position control via masking without compromising generation quality. This highlights the robustness and generalization capability of our model.

\subsection{\textit{Human Evaluation}}

\begin{table}[htbp]
\centering
\renewcommand{\arraystretch}{1.2}
\begin{tabular}{lccc}
\toprule
\textbf{Name} & \textbf{FID ↓} & \textbf{CLIP-I ↑} & \textbf{CLIP-T ↑} \\
\midrule
StoryGen(base)\_O   & 175 & 0.72 & 0.26 \\
StoryDALLE          & 168 & 0.71 & 0.24 \\
ARLDM               & 200 & 0.75 & 0.20 \\
NAME-A\_A           & 176 & 0.69 & 0.23 \\
StoryGen(base)\_A   & 195 & 0.70 & 0.26 \\
\textbf{NAME (ours)}& \textbf{152} & \textbf{0.72} & \textbf{0.24} \\
\bottomrule
\end{tabular}
\caption{Comparison of different models based on FID, CLIP-I, and CLIP-T. StoryGen(base)\_O refers to StoryGen trained on our dataset, while StoryGen(base)\_A is trained on the official full dataset, StorySalon. NAME-A\_A denotes our model variant without the Aug module, trained on the full dataset. StoryDALLE, ARLDM, and NAME are all trained on our dataset.}
\label{comparision_7:3}
\end{table}

In line with our objective to alleviate reading pressure for both researchers and participants during narrative inquiry member checking, human evaluation plays a crucial role. Our aim is to support more reasonable and coherent story generation while minimizing potential negative effects on participants.
Reading pressure in a narrative is influenced by multiple factors. To comprehensively assess our model's performance, we conduct human evaluations across the following key dimensions: character consistency, text-image consistency, reading pressure, character positioning consistency, and detail consistency. The evaluation results are summarized in Table~\ref{tab:consistency_comparison}.

\begin{table*}[htbp]
\centering
\renewcommand{\arraystretch}{1.2}
\begin{adjustbox}{width=\textwidth}
\begin{tabular}{lccccc}
\toprule
\textbf{Model} & \textbf{Character Consistency ↑} & \textbf{Text-Image Consistency ↑} & \textbf{Position Consistency ↑} & \textbf{Detail Consistency ↑} & \textbf{Reading Effort ↓} \\
\midrule
Pure Text       & \textbackslash{} & \textbackslash{} & \textbackslash{} & \textbackslash{} & 2.20 \\
ARLDM           & 1.65             & 2.00             & 2.67             & 1.90             & 1.33 \\
StoryDALLE      & 1.63             & 2.61             & 2.13             & 1.62             & 1.83 \\
StoryGen (Base) & 2.38             & 3.29             & 2.93             & 1.98             & 2.00 \\
\textbf{NAME (Ours)} & \textbf{3.12} & \textbf{3.52} & \textbf{4.13} & \textbf{3.33} & \textbf{1.16} \\
\bottomrule
\end{tabular}
\end{adjustbox}
\caption{Comparison of different models based on human evaluation of Character Consistency, Text-Image Consistency, Position Consistency, Detail Consistency, and Reading Effort.}
\label{tab:consistency_comparison}
\end{table*}

Notable enhancements in character positioning and detail consistency suggest that our model more reliably maintains visual continuity and preserves story-specific elements across scenes. This results in clearer and more logically consistent story progression, reducing the likelihood of confusion or misinterpretation.
Furthermore, our model achieves the lowest reading pressure score (1.16), indicating that the generated content is easier to process and understand. Lower reading pressure not only facilitates comprehension for participants during member checking but also aids researchers in analysis and evaluation.
Taken together, these findings suggest that our model generates visual stories that are not only more structured and logically coherent but also more accessible and cognitively efficient. By minimizing unnecessary complexity and promoting clearer storytelling, our approach enhances the effectiveness and reduces understanding pressure of the member checking process.

\section{Dataset}
In this section, we reflect on both the strengths and current limitations of our approach to story generation.

The StorySalon dataset, which serves as the foundation for our research, is a well-curated and authoritative resource, originally designed for multimodal story understanding and generation. Compared to many crowdsourced or synthetic datasets, it poses lower copyright and annotation risks, making it a safe and reliable choice for early-stage exploratory research. Furthermore, its narrative diversity and rich visual content make it particularly well-suited for studying character-grounded story generation, even though it was not explicitly constructed for character position control tasks. For our purposes, StorySalon strikes a practical balance between quality, accessibility, and research suitability.

However, the dataset presents several inherent limitations that fundamentally shaped our data construction methodology, particularly the lack of pre-existing segmentation masks and the high prevalence of multi-character scenes. These factors pose significant challenges for tasks requiring precise spatial control over individual entities. To overcome these issues, we developed a multi-stage pipeline that integrates automated processing with comprehensive human oversight:
(i) Initial segmentation: We employed SAM2 \cite{ravi2024sam2} to generate coarse segmentation masks, serving as a starting point to delineate potential character regions.
(ii) Manual selection and refinement: Human annotators then carried out an extensive curation process-including identifying single-character images, refining and correcting masks, eliminating irrelevant or ambiguous regions, and ensuring semantic alignment between image content and textual descriptions.
(iii) Quality assurance: Multiple rounds of manual review and cross-validation were conducted to uphold consistency, accuracy, and overall dataset integrity.
This meticulous pipeline, executed over the course of approximately one month and involving significant manual effort, resulted in a substantial size reduction-ultimately retaining just 0.96 percent of the original 124,918 images. Yet this curated subset achieves markedly improved spatial fidelity and semantic alignment, offering a high-quality foundation for downstream tasks that demand fine-grained character control.

Despite its reduced scale, we believe the resulting subset is a valuable step toward building structured datasets for controllable story generation. It enables focused experimentation and provides a strong starting point for future work. We plan to release this refined dataset publicly, with the hope that it will support the community in developing more advanced models and inspire the creation of richer, more purpose-built datasets in the future.

\section{Narrative Inquiry Based Evaluation Metrics}

Narrative inquiry focuses on understanding participants' lived experiences and how they make sense of them. Building on this perspective, to evaluate the impact of our model on participants, we not only propose three distinct evaluation metrics-Character Visual Consistency (CVC), Spatial Narrative Consistency (SNC), and Character Form Consistency (CFC)-but also integrate them into a unified composite metric to provide a more comprehensive assessment. The results are shown in Table~\ref{metrics_transposed}.

\begin{table*}[htbp]
\centering
\renewcommand{\arraystretch}{1.2}
\setlength{\tabcolsep}{10pt} 
\begin{tabular}{c|c|cc|ccc|c}
\toprule
\multirow{2.5}{*}{\textbf{Model}} & 
\textbf{CVC} & 
\multicolumn{2}{c|}{\textbf{SNC}} & 
\multicolumn{3}{c|}{\textbf{CFC}} & 
\multirow{2.5}{*}{\textbf{Overall}} \\
\cmidrule(lr){2-2} \cmidrule(lr){3-4} \cmidrule(lr){5-7}
 & \textbf{CN ↓} & \textbf{SR ↑} & \textbf{LA ↑} & \textbf{BDP ↓} & \textbf{MC ↓} & \textbf{ADS ↓} & \\

\midrule
ARLDM      & 184 & 0.14 & 0.08 & 238 & 212 & 91 & 1.90 \\
StoryDALLE & 190 & 0.17 & 0.11 & 101 & 93  & 49 & 2.68 \\
StoryGen-O & 148 & 0.16 & 0.24 & 101 & 193 & 45 & 2.66 \\
StoryGen-A & 160 & 0.11 & 0.18 & 131 & 119 & 54 & 2.57 \\
NAME-A\_A  & 151 & 0.19 & 0.29 & 119 & 106 & 45 & 2.89 \\

\textbf{NAME (Ours)} & \textbf{132} & \textbf{0.49} & \textbf{0.34} & \textbf{80} & \textbf{69} & \textbf{23} & \textbf{3.62} \\

\bottomrule
\end{tabular}
\caption{Comparison of proposed metrics across multiple dimensions, including Character Visual Consistency (CVC), Spatial Narrative Consistency (SNC), and Character Form Consistency (CFC), evaluated in terms of Credibility and Naturalism (CN), Smaller Regions (SR), Localization Accuracy (LA), Boundary Points (BDP), Mean-Case (MC), and Average Deviation along the Surface (ADS). StoryGen(base)\_O refers to StoryGen trained on our dataset, while StoryGen(base)\_A is trained on the official full dataset, StorySalon. NAME-A\_A denotes our model variant without the Aug module, trained on the full dataset. StoryDALLE, ARLDM, and NAME are all trained on our dataset.}
\label{metrics_transposed}
\end{table*}

\subsection{Character Visual Consistency}

Focusing on Character Visual Consistency, we recognize that visual coherence plays a pivotal role in shaping how participants perceive and connect with narrative agents. A high level of visual consistency enhances the authenticity and perceived credibility of characters, which in turn facilitates deeper identification and emotional engagement \cite{Hosokawa2024}. Existing research has shown that when a character's visual representation aligns with the viewer's internalized mental model, it can foster a temporary suspension of self-awareness, allowing for a more fluid emotional connection with the character \cite{doi:10.1177/1046878120926694}. In this context, consistency is not merely a stylistic preference but a perceptual anchor-supporting a natural and uninterrupted sense of presence.

Realistic character appearances serve as key entry points for participants to engage meaningfully with narrative environments. Visually coherent representations help maintain the internal logic of the story world, supporting empathetic engagement and narrative continuity. Conversely, inconsistencies in visual realism is likely to disrupt this balance to some extent, introducing cognitive dissonance that can diminish immersion and affective resonance \cite{XinyueZhang2023} \cite{10.3389/fpsyg.2017.01190}. Consequently, Character Visual Consistency serves as a core evaluative dimension within our framework.

In order to assess Character Visual Consistency, we segment both generated character images and corresponding reference regions using predefined masks, and compute the Fréchet Inception Distance (FID) \cite{Seitzer2020FID} as an indicator of visual similarity. The FID provides an interpretable metric for assessing the credibility and naturalism (CN) of a character's appearance. Lower FID scores suggest higher degrees of perceptual alignment, which support a more continuous and emotionally resonant experience-thereby preserving the narrative flow and enhancing the psychological plausibility of character interactions.

\subsection{Spatial Narrative Consistency}

Spatial Narrative Consistency captures the extent to which a character's spatial placement within a scene aligns with the implicit logic and expectations of the narrative. High spatial consistency ensures the coherence of the story world's spatial organization, which is critical for sustaining narrative flow and perceptual believability \cite{Magliano2016}. A well-maintained sense of spatial presence-the felt experience of being situated within the story space-can profoundly shape the immersion depth. By contrast, characters that appear misaligned, floating, or positioned implausibly may interrupt spatial continuity, undermining the viewer's internal mapping of the environment and weakening the narrative's overall coherence \cite{XinyueZhang2023}.

Precise spatial placement enables participants to track narrative events more intuitively, reinforcing spatial memory and facilitating comprehension of character actions and interactions \cite{soni2023groundingcharactersplacesnarrative}. When this consistency is compromised, spatial reasoning can become effortful, increasing cognitive load and diverting attention away from the story itself \cite{articleRapp}. Such disruptions can fragment the immersive experience, compelling the viewer to recalibrate their mental model of the scene-often at the cost of emotional continuity and narrative engagement.

In order to quantify Spatial Narrative Consistency, we employ YOLOv8 \cite{yolov8} for object detection and SAM2 \cite{ravi2024sam2} for segmentation, generating masks that localize character positions within each scene. 

To evaluate the overall overlap between predicted and reference regions and obtain a robust measure of localization accuracy (LA), we compute mIoU \cite{Everingham2010}. Further emphasizing spatial overlap and account for sensitivity to smaller regions (SR), we calculate the Dice coefficient \cite{shamir2019continuousdicecoefficientmethod}.  

Higher scores in these metrics suggest tighter alignment with spatial expectations, minimizing perceptual disruptions and preserving a fluid and uninterrupted narrative experience.

\subsection{Character Form Consistency}

Character Form Consistency assesses the fidelity of a generated character's shape, including body configuration, contours, and morphological structure. High form consistency reflects a closer alignment between the generated character and real-world references in terms of body proportions and naturalistic dynamics. These structural attributes are integral to affective realism-the perception that emotional expression emerges from credible visual stimuli \cite{doi:10.1177/1046878120926694}. When character posture and shape are rendered with accuracy and nuance, the resulting figure appears more lifelike and psychologically coherent, thereby strengthening emotional resonance. Conversely, distortions in form-whether exaggerated or subtly unnatural-may introduce perceptual dissonance that diminishes believability and interrupts viewer engagement.

Maintaining consistent morphological detail supports the expressive clarity of character behavior, making emotional states and narrative intentions more legible through posture and physical nuance \cite{10.1007/978-981-16-0041-8_26}. Subtle variations in form can convey distinct personality traits or narrative tension, while degradation in shape quality may compromise expressiveness and reduce emotional salience. This sensitivity is particularly salient in hyper-realistic contexts, where even minor deviations can disrupt immersion or evoke discomfort \cite{doi:10.1177/1046878120926694}.

However, structural differences pertaining to morphological detail consistency-such as limb length and proportional distribution-are often inadequately captured by conventional region-overlap metrics such as mIoU or Dice. To precisely quantify these structural-level morphological discrepancies, we propose an evaluation framework based on boundary-space deviations. Specifically, character masks are first subjected to edge detection using the OpenCV-Python library to extract fine-grained contours. Subsequently, morphological similarity is assessed using three complementary metrics:

(i) To capture the maximum deviation between boundary points (BDP), we compute the Hausdorff Distance \cite{232073}; (ii) To provide a more stable mean-case (MC) evaluation, we compute the Modified Hausdorff Distance (ModHausdorff) \cite{576361}; (iii) To quantify the average deviation along the surface (ADS), we use the Average Surface Distance (ASD) \cite{HEIMANN2009543}.

Lower values across these metrics indicate stronger structural alignment, suggesting a greater likelihood of sustained emotional engagement and narrative coherence.

\subsection{Integration}

To facilitate a unified evaluation, we integrate the six metrics across the three aforementioned dimensions. Specifically, for Dice and mIoU, we retain their original values. In contrast, since FID, Hausdorff distance, modified Hausdorff distance (ModHausdorff), and average surface distance (ASD) all range over $[0,\infty)$ and favor smaller values, we employ a monotonically decreasing transformation function $f\colon [0,\infty)\to[0,1]$. Our desiderata for $f$ are as follows:

(i) Domain: $x\ge0$.

(ii) Codomain: $0\le f(x)\le1$.

(iii) Strictly decreasing on its domain.

(iv) Uniform rate of decrease over the interval of interest.

(v) Continuous and uniformly continuous on $[0,\infty)$.

(vi) Everywhere differentiable on $[0,\infty)$.

The domain choice stems from the natural range of the four metrics, while the codomain $[0,1]$ aligns them with Dice and mIoU for subsequent aggregation. Monotonicity ensures that smaller metric values yield larger transformed scores. To achieve a roughly constant sensitivity across the operative range-practically up to $x=400$ for all four metrics-we impose a near-uniform derivative magnitude. Continuity and uniform continuity guarantee stability and preclude abrupt score fluctuations, whereas differentiability facilitates analytical tractability. We compared several candidate functions \eqref{eq:f1}, \eqref{eq:f2}, \eqref{eq:f3} and \eqref{eq:f4}:

\begin{figure}[htbp]
  \centering
  \begin{subfigure}[c]{0.35\textwidth}
    \raggedright
    \begin{align}
    f_1(x) &= \exp\left(-\frac{x}{200}\right) \label{eq:f1}, \\
    f_2(x) &= \frac{1}{x+1} \label{eq:f2}, \\
    f_3(x) &= \frac{1}{\sqrt{x+1}} \label{eq:f3}, \\
    f_4(x) &= -\frac{x}{400} + 1   \label{eq:f4},
    \end{align}
  \end{subfigure}%
  \hfill
  \begin{subfigure}[c]{0.6\textwidth}
    \raggedright
    
    \begin{tikzpicture}
    \begin{axis}[
        width=4cm,
        height=3cm,
        xlabel={$x$},
        ylabel={$f(x)$},
        title={$f_1$},
        xmin=0, xmax=500,
        ymin=0, ymax=1.1,
        grid=major,
        samples=200,
        domain=0:500
    ]
    \addplot[blue, thick, smooth] {exp(-x/200)};
    \end{axis}
    \end{tikzpicture}
    \begin{tikzpicture}
    \begin{axis}[
        width=4cm,
        height=3cm,
        xlabel={$x$},
        ylabel={$f(x)$},
        title={$f_2$},
        xmin=0, xmax=500,
        ymin=0, ymax=1.1,
        grid=major,
        samples=200,
        domain=0:500
    ]
    \addplot[red, thick, smooth] {1/(x+1)};
    \end{axis}
    \end{tikzpicture}
    
    \vspace{0.3cm}

    \begin{tikzpicture}
    \begin{axis}[
        width=4cm,
        height=3cm,
        xlabel={$x$},
        ylabel={$f(x)$},
        title={$f_3$},
        xmin=0, xmax=500,
        ymin=0, ymax=1.1,
        grid=major,
        samples=200,
        domain=0:500
    ]
    \addplot[green!60!black, thick, smooth] {1/sqrt(x+1)};
    \end{axis}
    \end{tikzpicture}
    \begin{tikzpicture}
    \begin{axis}[
        width=4cm,
        height=3cm,
        xlabel={$x$},
        ylabel={$f(x)$},
        title={$f_4$},
        xmin=0, xmax=500,
        ymin=0, ymax=1.1,
        grid=major,
        samples=200,
        domain=0:500
    ]
    \addplot[orange, thick, smooth] {-x/400 + 1};
    \end{axis}
    \end{tikzpicture}
  \end{subfigure}
  \caption{Function definitions and their graphical representations. The \textcolor{blue}{blue line} represents $f_1(x) = e^{-x/200}$, the \textcolor{red}{red line} corresponds to $f_2(x) = \frac{1}{x+1}$, the \textcolor{green!60!black}{green line} depicts $f_3(x) = (x+1)^{-\tfrac{1}{2}}$, and the \textcolor{orange}{orange line} shows $f_4(x) = -\frac{x}{400} + 1$.}
  \label{function}
\end{figure}
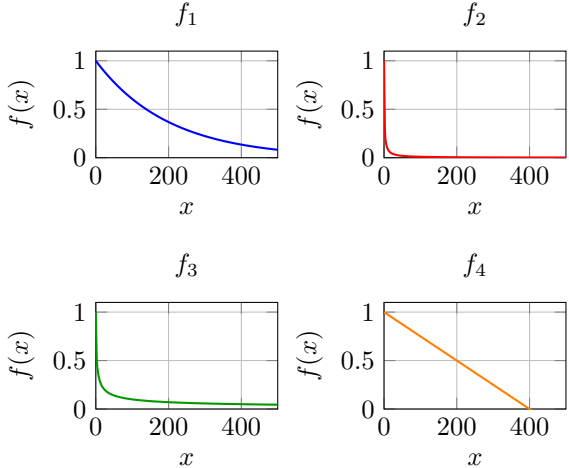

Fig.~\ref{function} illustrates the candidate functions under consideration. Among them, only \( f_1(x) \) and \( f_4(x) \) satisfy all of our predefined criteria. In contrast, \( f_2(x) \) and \( f_3(x) \) exhibit a steep decline when \( x < 100 \), rendering the function values nearly insensitive to changes in the input when \( x > 100 \). This insensitivity impairs the ability to accurately capture variations in the corresponding evaluation metrics. 

Although \( f_4(x) \) is linear and exhibits uniform variation across its domain, it lacks the desired sensitivity near optimal values for certain metrics-namely, FID, Hausdorff distance, modified Hausdorff distance (ModHausdorff), and average surface distance (ASD). These metrics are typically more difficult to optimize as they approach their ideal (i.e., near-zero) values. Therefore, we posit that the absolute value of the derivative should increase as the metric approaches zero, thereby emphasizing improvements near the optimal range.

Accordingly, we adopt \( f_1(x) \) for transforming FID, Hausdorff, ModHausdorff, and ASD. The transformed values are subsequently aggregated with Dice and mIoU scores to compute the overall evaluation score.

\section{Ablation}
\label{Ablation}

In this section, we present ablation studies in three stages. First, we evaluate the individual contribution of each integrated module to the overall performance of our model. Next, we investigate how the order of training affects the overall performance of the model. Finally, we perform an ablation study on our proposed augmentation module. For each part, we report two sets of evaluation metrics: standard metrics, including FID, CLIP-I, and CLIP-T, and our proposed metric tailored for narrative understanding.

\subsection{Module}
\begin{figure}[htbp]
  \centering
  \includegraphics[
    width=0.48\textwidth,
  ]{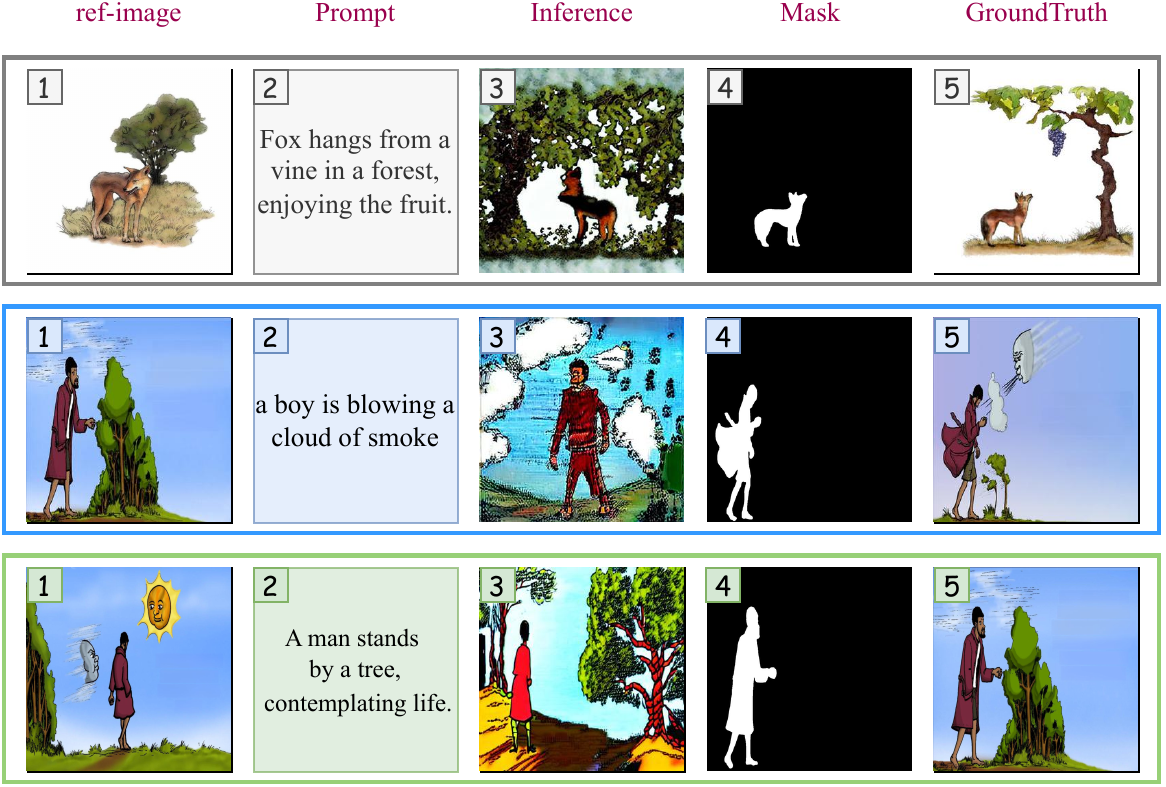}
  \caption{Visualization on model, from top to bottom: 
\textcolor{graybox}{Base}, 
\textcolor{bluebox}{Base and Aug}, and 
\textcolor{greenbox}{Base and Position control}.}
  \label{modulevisualization}
\end{figure}

Our model comprises three components: the base model, the edited text-mask cross-attention module, and the mask-augmented module. To evaluate the effectiveness of each module, we conduct a series of ablation studies. To ensure the rigor and reliability of our experimental results, we consider two data partitioning strategies: (1) the default configuration of our base model StoryGen, using 95 percent of the data for training and 5 percent for testing, and (2) a widely adopted split allocating 70 percent for training and 30 percent for testing. The corresponding results are reported in Table~\ref{module ablation on standard metrics}. In addition, we assess the impact on member checking based on our proposed metrics under the 70/30 split in Table~\ref{Module ablation Study for NAME on our proposed metrics}. The visualization of the results is presented in Fig.~\ref{modulevisualization}.

\begin{table*}[htbp]
\centering
\begin{tabular}{ccc|ccc|ccc}
\toprule
\multicolumn{3}{c|}{module settings} & \multicolumn{3}{c|}{19:1 metrics} & \multicolumn{3}{c}{7:3 metrics} \\
\cmidrule(r){1-3} \cmidrule(r){4-6} \cmidrule(r){7-9}
Base & Aug method & position control & FID ↓ & CLIP-I ↑ & CLIP-T ↑ & FID ↓ & CLIP-I ↑ & CLIP-T ↑ \\
\midrule
\checkmark &           &             & 96    & 0.75  & 0.25  & 175 & 0.72   & 0.26  \\
\checkmark & \checkmark &             & 63    & 0.74   & 0.27  & 169 & 0.73   & 0.27  \\
\checkmark &           & \checkmark  & 57    & 0.70   & 0.24  & 166 & 0.71   & 0.23  \\
\checkmark & \checkmark & \checkmark  & \textbf{49} & \textbf{0.72} & \textbf{0.25} & \textbf{152} & \textbf{0.72} & \textbf{0.24} \\
\bottomrule
\end{tabular}
\caption{Ablation study on our baseline, Aug module, and position control module on FID, CLIP-I and CLIP-T with two dataset split methods 19:1 and 7:3.}
\label{module ablation on standard metrics}
\end{table*}

\begin{table*}[htbp]
\centering
\renewcommand{\arraystretch}{1.2}
\setlength{\tabcolsep}{10pt}
\begin{adjustbox}{width=\textwidth}
\begin{tabular}{ccc|c|cc|ccc|c}
\toprule
\multicolumn{3}{c|}{\textbf{Module Settings}} & 
\textbf{CVC} & 
\multicolumn{2}{c|}{\textbf{SNC}} & 
\multicolumn{3}{c|}{\textbf{CFC}} & 
\multirow{2.5}{*}{\textbf{Overall}} \\
\cmidrule(r){1-3} \cmidrule(r){4-4} \cmidrule(r){5-6} \cmidrule(r){7-9}
\textbf{Base} & \textbf{Aug Method} & \textbf{Position Control} & 
\textbf{CN ↓} & 
\textbf{SR ↑} & \textbf{LA ↑} & 
\textbf{BDP ↓} & \textbf{MC ↓} & \textbf{ADS ↓} & \\
\midrule
\checkmark &             &               &147& 0.16 &  0.25  & 108   &     100    &    48  & 2.87 \\
\checkmark & \checkmark  &               &147&0.18 &  0.28 & 107  &  97    &    45 & 2.94 \\
\checkmark &             & \checkmark    &135& 0.20 &  0.30 & 92     &   84    &    34 & 3.14 \\
\checkmark & \checkmark  & \checkmark    &\textbf{132}& \textbf{0.49} & \textbf{0.34} & \textbf{80} & \textbf{69} & \textbf{23} & \textbf{3.62} \\
\bottomrule
\end{tabular}
\end{adjustbox}
\caption{Ablation study on our baseline, Aug module, and position control module, evaluated across multiple metrics, including Character Visual Consistency (CVC), Spatial Narrative Consistency (SNC), and Character Form Consistency (CFC), with evaluation dimensions covering Credibility and Naturalism (CN), Smaller Regions (SR), Localization Accuracy (LA), Boundary Points (BDP), Mean-Case (MC), and Average Deviation along the Surface (ADS).}
\label{Module ablation Study for NAME on our proposed metrics}
\end{table*}

For standard metrics ablation, we observe that: (i) The results indicate that, for the same dataset, different data splitting strategies can lead to significant variations in model performance.  In particular, using 95 percent of the data for training yields notably better performance compared to using only 70 percent.
(ii) When comparing our baseline model with its augmented variant (baseline+Aug), both data splitting strategies yield comparable outcomes.  Our method demonstrates substantial improvement in terms of FID.  For the CLIP-I metric, one strategy shows a slight increase while the other shows a slight decrease, both within 1 percent, indicating a negligible difference.  
In terms of CLIP-T, there is a marginal improvement.
(iii) Comparing the baseline model with the version enhanced by the mask control module (baseline+position), we observe a significant improvement in FID.  However, there is a slight decline in both CLIP-I and CLIP-T scores.
(iv) In comparisons between our proposed model and the baseline, both additional modules contribute positively to the FID score, resulting in a noticeable overall performance gain. 
Regarding CLIP-based metrics, one splitting strategy results in a comparable CLIP-I score with a slight drop in CLIP-T, while the other shows the reverse-CLIP-T remains stable with a slight decrease in CLIP-I.  These fluctuations are minor, suggesting that our model maintains competitive performance across both strategies.

In contrast, the ablation study based on our proposed metrics provides more detailed insights into the narrative consistency aspects of the model. Specifically, the augmentation module achieves performance comparable to the baseline on the Character Visual Consistency (CVC) metric, with only minor improvements observed in Spatial Narrative Consistency (SNC) and Character Form Consistency (CFC). 

As a result, the overall member checking score under our proposed evaluation framework shows only a slight increase. These findings suggest that the primary contribution of the augmentation module lies in enhancing overall image quality, rather than in narrative or character consistency.

In comparison, the position control module delivers more noticeable improvements across all three dimensions-CVC, SNC, and CFC-suggesting its stronger capacity to preserve visual and spatial coherence throughout the narrative. As a result, it leads to a more substantial enhancement in the member checking.

When both the augmentation and position control modules are integrated, their complementary strengths contribute to significant improvements across all proposed metrics. This combination achieves the highest member checking score of 3.62, highlighting the effectiveness of our full model in maintaining narrative consistency and character coherence.

\subsection{Training order}
\label{Attention layer}

\begin{figure*}[htbp]
  \centering
  \begin{subfigure}[c]{0.48\textwidth}
    \centering
  \includegraphics[
    width=\textwidth,
  ]{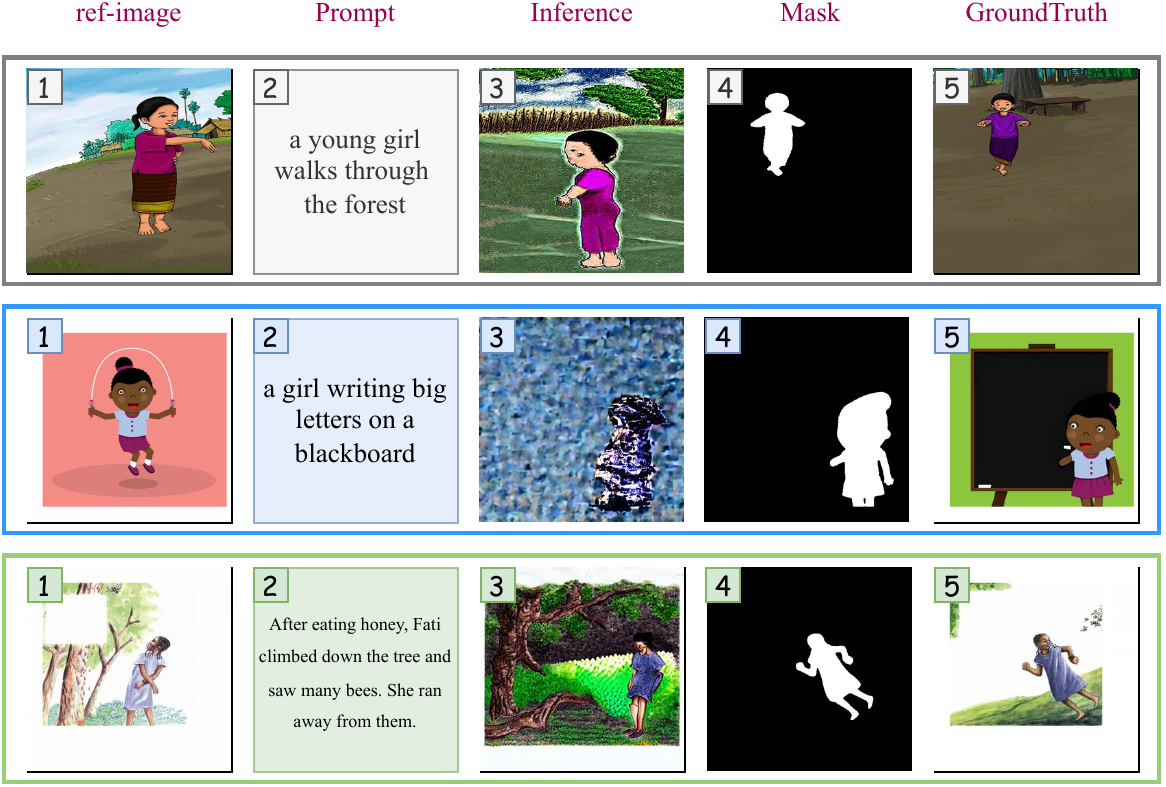}
  \caption{Visualization on model, from top to bottom: 
\textcolor{graybox}{GeR}, 
\textcolor{bluebox}{GsGe}, and 
\textcolor{greenbox}{GsGeR}.}
  \label{seqvisualization}
  \end{subfigure}
  \hfill
  \begin{subfigure}[c]{0.48\textwidth}
    \centering
    \renewcommand{\arraystretch}{1.2}
\begin{tabular}{lccc}
\toprule
\textbf{Name} & \textbf{FID ↓} & \textbf{CLIP-I ↑} & \textbf{CLIP-T ↑} \\
\midrule
GeR   & 160  & 0.71 & 0.23 \\ 
GsGe  & 303 & 0.60 & 0.21  \\
GsGeR & 167  &   0.70  &  0.24 \\
\textbf{GsRGe}& \textbf{152} & \textbf{0.72} & \textbf{0.24} \\
\bottomrule
\end{tabular}
\caption{Ablation study on different feature integration sequences:(i) GeR: The model first extracts geometric features, followed by joint refinement of semantic and geometric representations.(ii) GsGe: The model first encodes global semantic information, followed by the integration of spatial and character-level geometric features.(iii) GsGeR: The model begins with global semantic encoding, then incorporates geometric features, and finally applies joint refinement.(iv) GsRGe: The model first captures global semantic information, followed by semantic refinement to enhance contextual understanding, and finally integrates geometric features. Models are evaluated on FID, CLIP-I, and CLIP-T.}
    \label{Sequence}
  \end{subfigure}
  \caption{Combined figure showing visualization and ablation results}
  \label{Ordercombined}
\end{figure*}

Our model incorporates three distinct attention layers: Self-Attention Layer, Text-Mask Attention Layer, and Image Attention Layer. The Self-Attention Layer is designed to capture global semantic information; the Text-Mask Attention Layer focuses on learning spatial and character-specific geometric information; and the Image Attention Layer refines both geometric and semantic representations. Accordingly, we explore the following training strategies:
(i) First learn geometric features, followed by a refinement of both semantic and geometric information, denoted as GeR.
(ii) First learn global semantic information, then spatial and character-level geometric features, denoted as GsGe.
(iii) Learn global semantics initially, followed by the integration of geometric features, and conclude with refinement of joint semantic and geometric information, denoted as GsGeR.
(iv) Learn global semantics first, then reinforce semantic understanding, and finally incorporate geometric features, denoted as GsRGe. The results are reported in terms of standard metrics (Table~\ref{sequs}) and our proposed evaluation metrics (Table~\ref{Sequence}). Corresponding visualizations are provided in Fig.~\ref{seqvisualization}.

From the perspective of standard evaluation metrics, our observations are as follows:
(i) GeR: This design enables a more balanced integration of visual and textual modalities, leading to solid overall performance.
(ii) GsGe: Without a refinement stage, the model captures spatial and character-level geometric features, but the outputs remain coarse, resulting in subpar performance.
(iii) GsGeR: Performing refinement after semantic encoding undermines generation quality. Moreover, the early fusion of geometric features tends to disrupt semantic consistency, further degrading overall performance.
(iv) GsRGe: This configuration improves both text generation quality and spatial layout control, achieving consistently superior results across evaluation metrics.

Based on our proposed evaluation metrics, we draw the following conclusions:
GeR, which lacks global semantic features but includes a refinement stage, performs slightly worse in Character Visual Consistency (CVC).
In contrast, GsGe, without refinement of global semantics, achieves the lowest CVC score. However, with the integration of geometric information, it performs relatively well in both Spatial Narrative Consistency (SNC) and Character Form Consistency (CFC).
GsGeR, which incorporates both semantic and geometric information, further improves CVC and maintains strong performance on SNC. Nevertheless, the early integration of geometric features may interfere with maintaining consistent character contours, leading to a decline in CFC.
Notably, the GsRGe strategy achieves the best results overall, outperforming the other strategies across all consistency dimensions and yielding the highest member checking score. We believe this is due to its more balanced training configuration, which integrates spatial guidance (Gs), role-level reasoning (R), and generative enhancement (Ge) in an order that effectively captures both positional and character-level coherence. This indicates that the design and sequencing of training components can significantly influence the model's ability to maintain narrative consistency.

\begin{table*}[htbp]
\centering
\renewcommand{\arraystretch}{1.2}
\setlength{\tabcolsep}{10pt} 
\begin{tabular}{c|c|cc|ccc|c}
\toprule
\multirow{2.5}{*}{\textbf{Model}} & 
\textbf{CVC} & 
\multicolumn{2}{c|}{\textbf{SNC}} & 
\multicolumn{3}{c|}{\textbf{CFC}} & 
\multirow{2.5}{*}{\textbf{Overall}} \\
\cmidrule(lr){2-2} \cmidrule(lr){3-4} \cmidrule(lr){5-7}
& \textbf{CN ↓} & \textbf{SR ↑} & \textbf{LA ↑} & \textbf{BDP ↓} & \textbf{MC ↓} & \textbf{ADS ↓} & \\

\midrule
GeR &143& 0.19 & 0.29 & 100 & 90  & 38 & 3.04 \\
GsGe  &194&  0.21 &  0.31 & 99  &  91 &  37 & 2.97 \\
GsGeR   &139&  0.21  & 0.31 & 110 &  101 &  47  & 2.99 \\
GsRGe       &\textbf{132}& \textbf{0.49} & \textbf{0.34} & \textbf{80} & \textbf{69} & \textbf{23} & \textbf{3.62} \\
\bottomrule
\end{tabular}
\caption{Ablation study on different feature integration sequences:(i) GeR: The model first extracts geometric features, followed by joint refinement of semantic and geometric representations.(ii) GsGe: The model first encodes global semantic information, followed by the integration of spatial and character-level geometric features.(iii) GsGeR: The model begins with global semantic encoding, then incorporates geometric features, and finally applies joint refinement.(iv) GsRGe: The model first captures global semantic information, followed by semantic refinement to enhance contextual understanding, and finally integrates geometric features. Models are evaluated on Character Visual Consistency (CVC), Spatial Narrative Consistency (SNC), and Character Form Consistency (CFC), with evaluation dimensions including Credibility and Naturalism (CN), Smaller Regions (SR), Localization Accuracy (LA), Boundary Points (BDP), Mean-Case (MC), and Average Deviation along the Surface (ADS).}
\label{sequs}
\end{table*}

\subsection{Augment Module Ablation}

\begin{figure*}[htbp]
  \centering
  \begin{subfigure}[c]{0.48\textwidth}
    \centering
    \includegraphics[
      width=\textwidth,
    ]{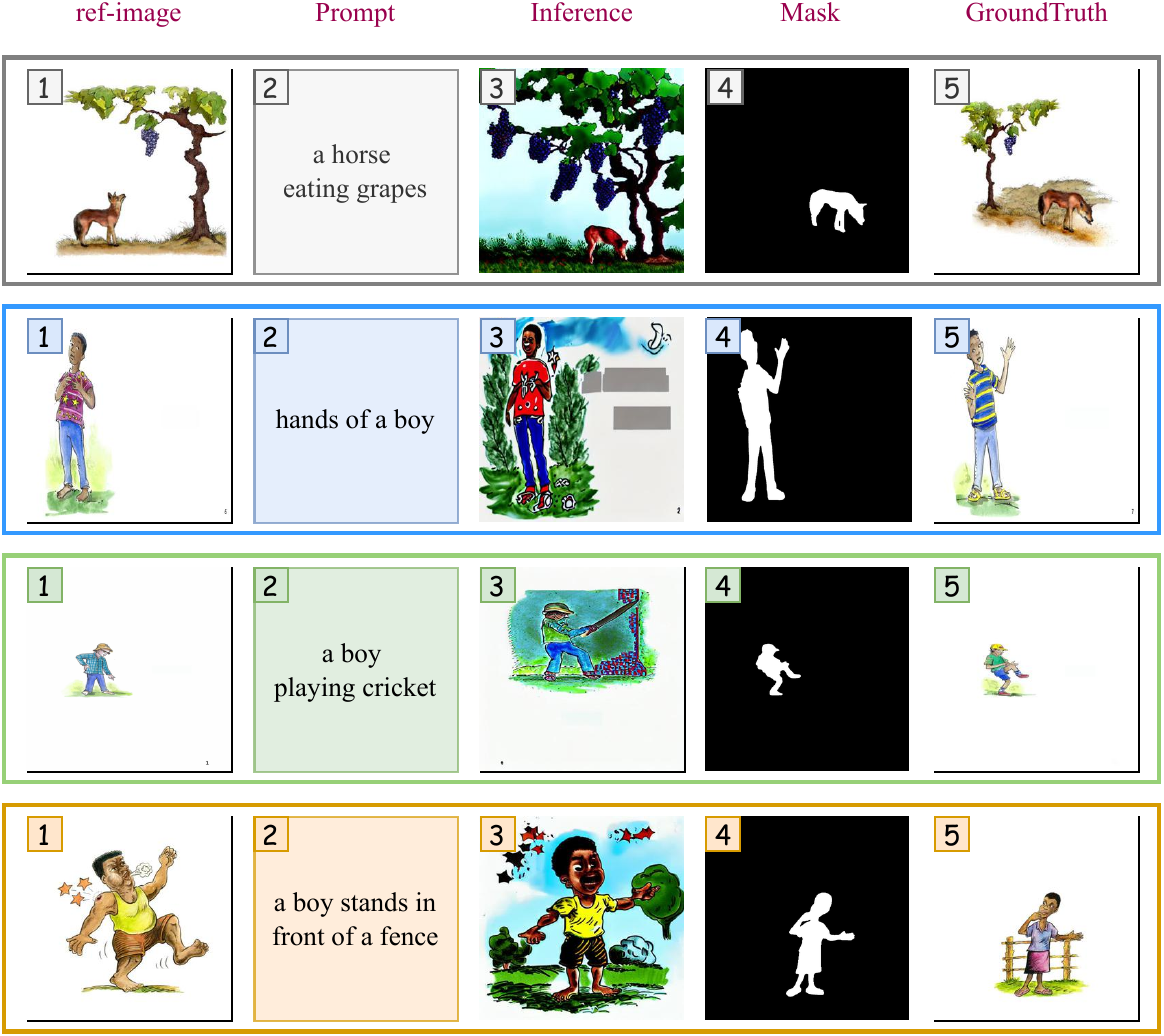}
    \caption{Visualization on model, from top to bottom: 
    \textcolor{graybox}{A-V-S-S-C}, 
    \textcolor{bluebox}{A-V-S-S}, 
    \textcolor{greenbox}{A-V-S}, and 
    \textcolor{orangebox}{A-V}.}
    \label{Augvisualization}
  \end{subfigure}
  \hfill
  \begin{subfigure}[c]{0.48\textwidth}
    \centering
    \renewcommand{\arraystretch}{1.2}
    \begin{tabular}{lccc}
      \toprule
      \textbf{Name} & \textbf{FID ↓} & \textbf{CLIP-I ↑} & \textbf{CLIP-T ↑} \\
      \midrule
      A-V-S-S-C &  159   &    0.71     &      0.24       \\
      A-V-S-S   &  154   &    0.72     &      0.24       \\
      A-V-S     &  163   &    0.69     &      0.23       \\
      A-V       &  165   &    0.70     &      0.23       \\
      \textbf{A} & \textbf{152} & \textbf{0.72} & \textbf{0.24} \\
      \bottomrule
    \end{tabular}
    \caption{Ablation study of the Augment module on FID, CLIP-I and CLIP-T. 
    ``A'' denotes the complete Augment model; 
    ``A-V'' removes the to vectors component; 
    ``A-V-S'' further removes the downsampling module; 
    ``A-V-S-S'' removes two downsampling steps; 
    and ``A-V-S-S-C'' additionally removes the convolution layer.}
    \label{Aug}
  \end{subfigure}
  \caption{Combined figure showing visualization and ablation results}
  \label{Augcombined}
\end{figure*}

In this section, we present an ablation study to evaluate the effectiveness of our proposed augmentation module. The results are illustrated in Table~\ref{augus} and Table~\ref{Aug}, and visualization results are shown in Fig.~\ref{Augvisualization}.

We evaluated the models using established benchmarks, including FID, CLIP-I, and CLIP-T. Under these conditions, the A-V-S-S-C underperformed relative to A-V-S-S, while A-V-S consistently yielded subpar results across all three metrics. Interestingly, the A-V exhibited moderate gains in CLIP-I and CLIP-T, despite continuing to lag behind in FID. Notably, the A model consistently achieved superior performance across all conventional evaluation metrics.  Beyond standard benchmarks, we further assessed model performance using our proposed evaluation metrics. While A-V-S-S-C and A-V-S-S demonstrated comparable overall performance, A-V-S-S-C performed relatively poorly on the SNC task, whereas A-V-S-S showed suboptimal outcomes on the CFC task.  In contrast, A-V-S consistently underachieved across all narrative-oriented tasks and evaluation dimensions. The A-V model demonstrated strong performance on both the SNC and CFC tasks but exhibited significant deficiencies in CVC. 
These performance disparities are further illustrated in Fig.~\ref{Augvisualization}, which provides a visual comparison of the models. To highlight the differences more clearly, we intentionally selected examples where the distinctions between models are particularly pronounced. These illustrative cases are intended to emphasize relative strengths, rather than to represent the entire performance distribution.

\begin{table*}[htbp]
\centering
\renewcommand{\arraystretch}{1.2}
\setlength{\tabcolsep}{10pt} 
\begin{tabular}{c|c|cc|ccc|c}
\toprule
\multirow{2.5}{*}{\textbf{Model}} & 
\textbf{CVC} & 
\multicolumn{2}{c|}{\textbf{SNC}} & 
\multicolumn{3}{c|}{\textbf{CFC}} & 
\multirow{2.5}{*}{\textbf{Overall}} \\
\cmidrule(lr){2-2} \cmidrule(lr){3-4} \cmidrule(lr){5-7}
& \textbf{CN ↓} & \textbf{SR ↑} & \textbf{LA ↑} & \textbf{BDP ↓} & \textbf{SMC ↓} & \textbf{ADS ↓} & \\

\midrule
A-V-S-S-C &139& 0.18 & 0.28 & 98& 87  & 37 & 3.05 \\
A-V-S-S  &136& 0.21 & 0.30 & 103 & 94  & 43 & 3.05 \\
A-V-S   &143& 0.17 & 0.26 & 116 & 104 & 47 & 2.86 \\
A-V    &140& 0.24 & 0.35 & 99  & 87  & 36 & 3,18 \\
A       &\textbf{132}& \textbf{0.49} & \textbf{0.34} & \textbf{80} & \textbf{69} & \textbf{23} & \textbf{3.62}\\
\bottomrule
\end{tabular}
\caption{Ablation study of the Augment module evaluated across multiple metrics, including Character Visual Consistency (CVC), Spatial Narrative Consistency (SNC), and Character Form Consistency (CFC), with evaluation dimensions covering Credibility and Naturalism (CN), Smaller Regions (SR), Localization Accuracy (LA), Boundary Points (BDP), Mean-Case (MC), and Average Deviation along the Surface (ADS).
``A'' denotes the complete Augment model; 
``A-V'' removes the to vectors component; 
``A-V-S'' further removes the downsampling module; 
``A-V-S-S'' removes two downsampling steps; 
and ``A-V-S-S-C'' additionally removes the convolution layer.}
\label{augus}
\end{table*}

\section{Conclusion}

Narrative inquiry, which focuses on understanding participants' lived experiences and personal stories, often demands a high level of accuracy in the materials used. To ensure this accuracy, particularly during the member checking phase, both researchers and participants are required to engage with extensive textual materials-an effort-intensive process that involves careful reading and reflection. This can place a significant burden of textual comprehension on both parties. To reduce the burden, our work introduces a controllable image generation framework focused on precise character positioning. By grounding image synthesis in spatially and semantically coherent prompts, we aim to reduce reliance on lengthy textual descriptions while improving visual clarity and alignment. This not only reduces the reading burden for researchers but also lowers participants' cognitive load during validation. Importantly, by enabling selective control over character inclusion and placement, our method allows for the generation of characters with more contextually appropriate shapes and positioning, thereby reducing unnecessary cognitive strain or psychological discomfort during member checking. We believe the first attempt in this field establishes a foundational step toward advancing research at the intersection of generative modeling and narrative inquiry in social sciences.

\textbf{Limitations:} As our approach is built upon diffusion models, certain limitations are inherently difficult to avoid. When discrepancies arise between the textual prompt and the visual cues in the reference image, the generation may be biased toward the reference, leading to semantic inconsistencies. At the same time, due to the lack of any publicly available dataset that offers coherent storylines with corresponding character masks, the promising capabilities of our model cannot be fully demonstrated.

\textbf{Future work:} In future work, we envision deeper integration of generative visual tools into qualitative and narrative inquiry workflows, with the goal of making interpretive processes more accessible and participant-friendly.

\bibliographystyle{IEEEtran}
\bibliography{reference.bib}

\end{document}